%% file: root.tex
\documentclass[letterpaper, 10 pt, conference]{ieeeconf}  

\IEEEoverridecommandlockouts                              

\usepackage{hyperref}

\PassOptionsToPackage{table}{xcolor} 
\usepackage[usenames,dvipsnames]{xcolor}
\usepackage{graphics} 
\usepackage{epsfig} 
\usepackage{mathptmx} 
\usepackage{times} 

\usepackage{stfloats} 
\usepackage{cuted}
\usepackage{capt-of}

\usepackage{cite}
\usepackage{amsmath,amssymb,amsfonts}
\usepackage{algorithm}
\usepackage{algpseudocode}
\usepackage{graphicx}
\usepackage{textcomp}
\usepackage{xcolor}
\usepackage{cleveref}
\usepackage{caption}
\usepackage{subfig}
\usepackage{float}      
\usepackage{booktabs}
\usepackage{multirow}
\newcommand{\best}[1]{\textbf{\textcolor{ForestGreen}{#1}}}
\newcommand{\secondbest}[1]{\textbf{\textcolor{LimeGreen}{#1}}}
\def\BibTeX{{\rm B\kern-.05em{\sc i\kern-.025em b}\kern-.08em
    T\kern-.1667em\lower.7ex\hbox{E}\kern-.125emX}}
\newcommand{\etal}{\textit{et al}.}

\title{\LARGE \bf
Robust Dynamic Object Detection in Cluttered Indoor Scenes via Learned Spatiotemporal Cues
}

\author{Albert Author$^{1}$ and Bernard D. Researcher$^{2}$
\thanks{*This work was not supported by any organization}
\thanks{$^{1}$Albert Author is with Faculty of Electrical Engineering, Mathematics and Computer Science,
        University of Twente, 7500 AE Enschede, The Netherlands
        {\tt\small albert.author@papercept.net}}%
\thanks{$^{2}$Bernard D. Researcheris with the Department of Electrical Engineering, Wright State University,
        Dayton, OH 45435, USA
        {\tt\small b.d.researcher@ieee.org}}%
}

\author{Juan Rached\textsuperscript{1}, Yixuan Jia\textsuperscript{1}, Kota Kondo\textsuperscript{1}, Jonathan P. How\textsuperscript{1}\\
\thanks{This work is supported in part by DSTA.}
\thanks{\textsuperscript{1}Massachusetts Institute of Technology, Cambridge, MA 02139, USA. \{\texttt{jrached, yixuany, kkondo, jhow}\}\texttt{@mit.edu}.}
}

\begin{document}

\pagestyle{plain} 

\maketitle
\thispagestyle{plain}

\begin{strip}
\vspace{-6.5em}
\centering
\includegraphics[width=\textwidth]{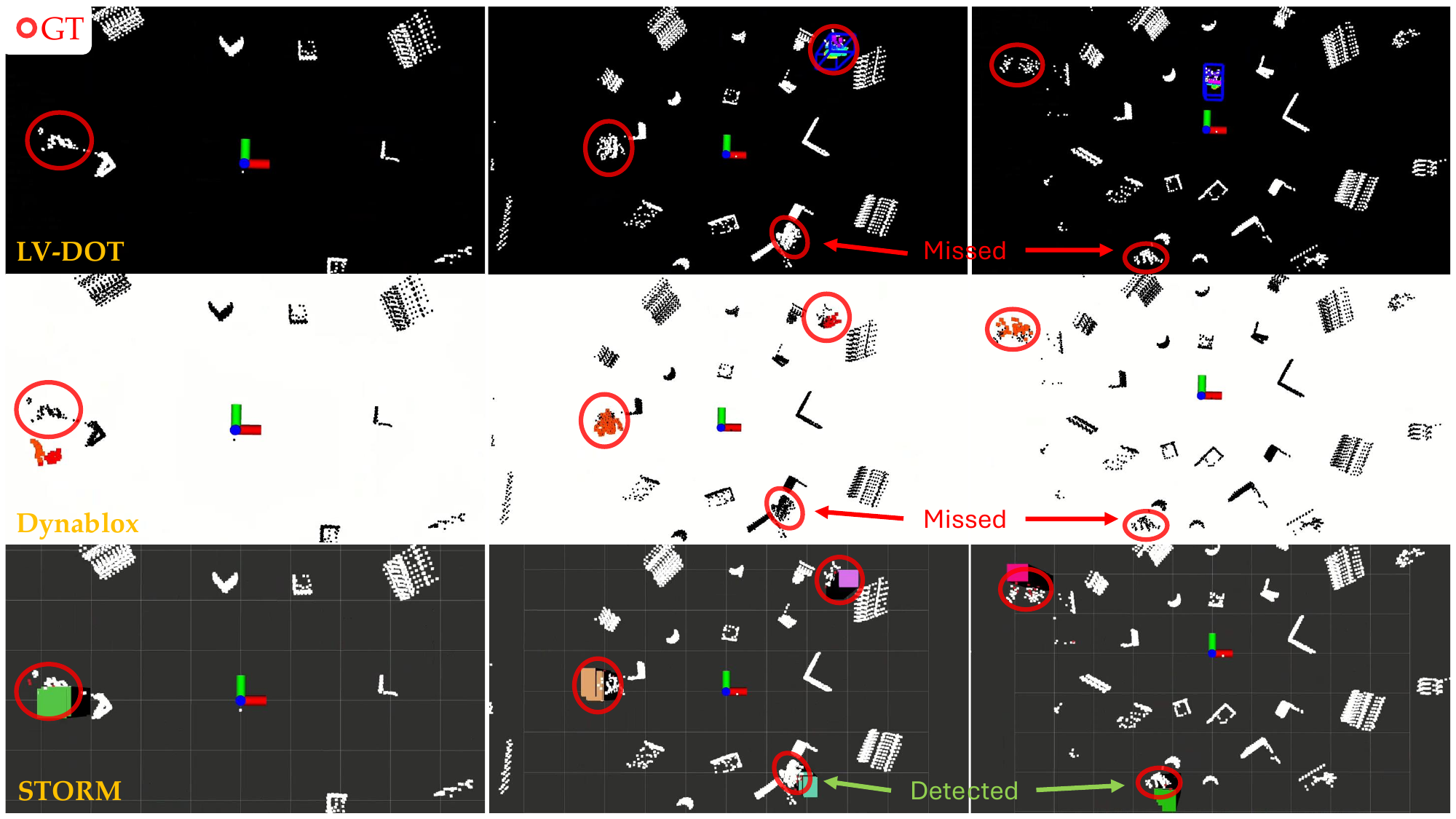}
\captionof{figure}{BEV of \texttt{LV-DOT} \cite{xu2025lv} (top row), \texttt{Dynablox} \cite{schmid2023dynablox} (middle row), and \texttt{STORM} (bottom row) across three experiments with increasing levels of clutter. \texttt{LV-DOT} struggles with proximity-induced false negatives in all experiments. \texttt{Dynablox} misses detections due to partial occlusions in columns 1 and 3 and due to proximity-induced false negatives in column 2. \texttt{STORM} detects all obstacles across all experiments.}
\label{fig:storm_exp}
\vspace{-0.8em}
\end{strip}
\vspace{-0.5em}



\begin{abstract}
Reliable dynamic object detection in cluttered environments remains a critical challenge for autonomous navigation. Purely geometric LiDAR pipelines that rely on clustering and heuristic filtering can miss dynamic obstacles when they move in close proximity to static structure or are only partially observed. Vision-augmented approaches can provide additional semantic cues, but are often limited by closed-set detectors and camera field-of-view constraints, reducing robustness to novel obstacles and out-of-frustum events. In this work, we present a LiDAR-only framework that fuses temporal occupancy-grid-based motion segmentation with a learned bird’s-eye-view (BEV) dynamic prior.
A fusion module prioritizes 3D detections when available, while using the learned dynamic grid to recover detections that would otherwise be lost due to proximity-induced false negatives. 
Experiments with motion-capture ground truth show our method achieves $\mathbf{28.67\%}$ higher recall and $\mathbf{18.50\%}$ higher F1 score than the state-of-the-art in substantially cluttered environments while maintaining comparable precision and position error. 
\end{abstract}


\section*{Supplementary Material}
\noindent\textbf{Video}: \href{https://youtu.be/\_ZHGz-BkQzg}{\url{https://youtu.be/\_ZHGz-BkQzg}} \\
\noindent\textbf{Code}: {To be released soon.}


\section{Introduction}
\input{introduction}

\section{Related Work}
\input{related_work}

\section{Method}
Our approach consists of two parallel detection pipelines. One uses the temporal occupancy information of voxels in an OGM and clustering to generate the positions of dynamic objects in 3D. This is covered in \Cref{subsec:togm}. The other compresses point cloud information from multiple scans into a sequence of pseudo-images that are fed into a Long Short Term Memory (LSTM) module. The embeddings produced by the LSTM are reconstructed into 2D dynamic grids where each square in the grid map represents either a static or dynamic column in 3D. These dynamic columns can be used to recover dynamic object detections that are lost in the clustering step of the 3D pipeline. This is covered in \Cref{subsec:gridnet}.

\subsection{Temporal Occupancy Grid Map Detector}\label{AA}
\label{subsec:togm}
\subsubsection{Moving Object Segmentation}
The MOS module of our 3D obstacle detection pipeline leverages the temporal information of each voxel in an OGM as in Dynablox \cite{schmid2023dynablox}. It works under the assumption that, if space is split into free and occupied regions, anything that moves into free space must be dynamic. This principle allows us to solve the MOS problem, and use dynamic point cloud segments as candidates for dynamic object detections. To distinguish between free and non-free space, the OGM keeps track of how long each voxel has been occupied and unoccupied. To that end, the internal state of each voxel in the grid map is augmented with variables $t_o^{(t)}$ and $t_u^{(t)}$, representing the last time that voxel was occupied and unoccupied respectively. The complete state of each voxel, $a^{(t)}$, is defined as follows, 
\begin{equation}
a^{(t)} = \{o^{(t)}, \space t_o^{(t)}, \space t_u^{(t)}, \space f^{(t)}\}\label{eq},
\end{equation}
where $o^{(t)} \in \{0, 1\}$, represents whether the voxel is occupied and $f^{(t)} \in \{0, 1\}$ whether it is considered free space at time $t$. 
At each timestep, if the voxel is occupied, the occupancy duration of that voxel, $t_o^{(t)}$, is updated. Once that number exceeds a threshold, $\tau_o$, the voxel is considered not free (i.e. $f^{(t)} = 0$). Similarly, if the voxel is unoccupied, the amount of time the voxel has been unoccupied, $t_u^{(t)}$, is updated. Once that value exceeds a threshold, $\tau_u$, the voxel is marked as free (i.e. $f^{(t)} = 1$). Any object that moves into free space, that is a voxel with fields $f^{(t)} = 1$ and $o^{(t)} = 1$, is considered dynamic and any voxel with $f^{(t)} = 0$ is considered static. Clustering and nearest-neighbors checks are applied to the dynamic subsets of the point cloud scans to compute position estimates and reduce potential false positives caused by noisy LiDAR readings. Unlike Dynablox, our OGM contains no hierarchical spatial structures as mapping potentially unbounded environments is not the focus of this work. In addition, our approach relies on cluster density heuristics, as opposed to point-count heuristics, to filter out noise, as we found it better handles artifacts introduced by large static surfaces.   

\begin{table*}[t]
\centering
\caption{Performance of methods on precision, recall, F1 score, and position error metrics across three distance thresholds on a set of experiments with 1-3 human dynamic obstacles. For each metric, the highest performing method is highlighted in \best{green} and the second highest in \secondbest{light green}. For precision, recall, and F1 score, higher is better. For position error, lower is better. \# Obs is the number of static obstacles in the room.}
\label{tab:human2}
\setlength{\tabcolsep}{3pt}
\renewcommand{\arraystretch}{1.1}
\scriptsize
\resizebox{\textwidth}{!}{
\begin{tabular}{cccccccccccccc}
\toprule
\multirow{2}{*}[-0.4em]{\# Obs} & \multirow{2}{*}[-0.4em]{Method}{}
& \multicolumn{4}{c}{DistThresh = 0.75m}
& \multicolumn{4}{c}{DistThresh = 0.5m}
& \multicolumn{4}{c}{DistThresh = 0.25m} \\
\cmidrule(lr){3-6}\cmidrule(lr){7-10}\cmidrule(lr){11-14}
&
& Precision & Recall & F1 & Pos Err ($m$)
& Precision & Recall & F1 & Pos Err ($m$)
& Precision & Recall & F1 & Pos Err ($m$) \\
\midrule
\multirow{5}{*}{4}
  & \texttt{LV-DOT}
& 0.96 & 0.75 & 0.84 & \best{0.14}
& 0.96 & 0.75 & 0.84 & \best{0.14}
& 0.96 & 0.74 & 0.83 & \best{0.13} \\
& \texttt{LV-DOT NC}
& 0.96 & 0.75 & 0.84 & \best{0.14}
& 0.96 & 0.75 & 0.84 & \best{0.14}
& 0.96 & 0.74 & 0.83 & \best{0.13} \\
& \texttt{Dynablox}
& \best{1.00} & \secondbest{0.91} & \secondbest{0.95} & 0.16
& \best{1.00} & \secondbest{0.91} & \secondbest{0.95} & 0.16
& \best{1.00} & \secondbest{0.90} & \secondbest{0.95} & \secondbest{0.15} \\
& \texttt{STORM NG}
& \secondbest{0.99} & \secondbest{0.91} & \secondbest{0.95} & \secondbest{0.15}
& \secondbest{0.99} & \secondbest{0.91} & \secondbest{0.95} & \secondbest{0.15}
& \secondbest{0.99} & 0.89 & 0.94 & \best{0.13} \\
& \texttt{STORM} (Ours)
& \secondbest{0.99} & \best{0.96} & \best{0.97} & 0.16
& \secondbest{0.99} & \best{0.96} & \best{0.97} & \secondbest{0.15}
& 0.98 & \best{0.96} & \best{0.97} & \best{0.13} \\
\midrule
\multirow{5}{*}{16}
  & \texttt{LV-DOT}
& 0.54 & 0.14 & 0.22 & \secondbest{0.16}
& 0.52 & 0.13 & 0.21 & \best{0.13}
& 0.51 & 0.12 & 0.20 & \best{0.12} \\
& \texttt{LV-DOT NC}
& \secondbest{0.95} & 0.05 & 0.10 & \best{0.13}
& \secondbest{0.95} & 0.05 & 0.10 & \best{0.13}
& \secondbest{0.94} & 0.05 & 0.10 & \best{0.12} \\
& \texttt{Dynablox}
& \best{1.00} & 0.60 & 0.75 & 0.17
& \best{1.00} & 0.60 & 0.75 & \secondbest{0.17}
& \best{1.00} & 0.56 & 0.72 & \secondbest{0.14} \\
& \texttt{STORM NG}
& \best{1.00} & \secondbest{0.66} & \secondbest{0.79} & 0.18
& \best{1.00} & \secondbest{0.65} & \secondbest{0.79} & \secondbest{0.17}
& \best{1.00} & \secondbest{0.61} & \secondbest{0.76} & \secondbest{0.14} \\
& \texttt{STORM} (Ours)
& \best{1.00} & \best{0.72} & \best{0.84} & 0.18
& \best{1.00} & \best{0.72} & \best{0.84} & \secondbest{0.17}
& \best{1.00} & \best{0.68} & \best{0.81} & \secondbest{0.14} \\
\midrule
\multirow{5}{*}{31}
  & \texttt{LV-DOT}
& 0.64 & 0.12 & 0.20 & \secondbest{0.17}
& 0.63 & 0.11 & 0.19 & \best{0.15}
& 0.61 & 0.11 & 0.18 & \best{0.13} \\
& \texttt{LV-DOT NC}
& 0.78 & 0.06 & 0.10 & \best{0.16}
& \secondbest{0.77} & 0.05 & 0.10 & \best{0.15}
& \secondbest{0.75} & 0.05 & 0.09 & \best{0.13} \\
& \texttt{Dynablox}
& \best{1.00} & 0.48 & 0.65 & 0.18
& \best{1.00} & 0.48 & 0.65 & \secondbest{0.18}
& \best{1.00} & 0.43 & 0.60 & \secondbest{0.15} \\
& \texttt{STORM NG}
& \best{1.00} & \secondbest{0.55} & \secondbest{0.71} & 0.18
& \best{1.00} & \secondbest{0.55} & \secondbest{0.71} & \secondbest{0.18}
& \best{1.00} & \secondbest{0.50} & \secondbest{0.67} & \secondbest{0.15} \\
& \texttt{STORM} (Ours)
& \best{1.00} & \best{0.60} & \best{0.75} & 0.18
& \best{1.00} & \best{0.60} & \best{0.75} & \secondbest{0.18}
& \best{1.00} & \best{0.55} & \best{0.71} & \secondbest{0.15} \\
\bottomrule
\end{tabular}}
\end{table*}

\subsubsection{Association and Estimation}
For each dynamic cluster, we construct a bounding box that contains all points belonging to that cluster and represent its position in 3D space using the cluster's centroid. Through our detection fusion procedure, discussed in more detail in \Cref{subsec:fusion}, cluster centroids are assigned an Extended Kalman Filter (EKF) estimate using greedy associations. To accurately capture a wide range of obstacle trajectories, including nonlinear motion, we use a constant acceleration Dubins plane model. We define each obstacle's state as, 
\begin{equation}
\mathbf{x} =
\begin{bmatrix}
x & y & z & \theta & \phi & v
\end{bmatrix}^\top
\label{eq:state}
\end{equation}
Where $x, y, z$ are the coordinates of the obstacle in $\mathbb{R}^3$, $\theta$ and $\phi$ are the heading and the climb angles respectively, and $v$ is the magnitude of the velocity of the object. The dynamics evolve as follows,
\begin{equation}
\label{equation:dynamics}
\dot{\mathbf{x}} =
\begin{bmatrix}
v \,c_\phi \,c_\theta, \; v \,c_\phi \,s_\theta, \; v \,s_\phi, \; \dot{\theta},\; \dot{\phi},\; a
\end{bmatrix} ^ T
\end{equation}
Where $c_\theta$ and $s_\theta$ are shorthand for $\cos\theta$ and $\sin\theta$, respectively. Following the standard EKF procedure, the right hand side of (\ref{equation:dynamics}) is discretized and linearized about the current best estimate to generate the state transition function $f(\cdot)$ and the state transition Jacobian $\mathbf{F}_k$. The dynamics are propagated using,
\[
\begin{aligned}
\hat{\mathbf{x}}_{k|k-1} &= f(\hat{\mathbf{x}}_{k-1|k-1}, \mathbf{u}_{k-1}) \\
\mathbf{P}_{k|k-1} &= \mathbf{F}_k \mathbf{P}_{k-1|k-1} \mathbf{F}_k^\top + \mathbf{Q}_k
\end{aligned}
\]
Here, $\hat{\mathbf{x}}_{k|k-1}$ and $\mathbf{P}_{k|k-1}$  are the predicted estimate and covariance at $k$ using measurements up to $k-1$. $\mathbf{Q}_k$ is the process noise covariance matrix. We follow the standard EKF update, 
\begin{align}
\mathbf{y}_k &= \mathbf{z}_k - h(\hat{\mathbf{x}}_{k|k-1}) \\
\mathbf{S}_k &= \mathbf{H}_k \mathbf{P}_{k|k-1} \mathbf{H}_k^\top + \mathbf{R}_k \\
\mathbf{K}_k &= \mathbf{P}_{k|k-1} \mathbf{H}_k^\top \mathbf{S}_k^{-1} \\
\hat{\mathbf{x}}_{k|k} &= \hat{\mathbf{x}}_{k|k-1} + \mathbf{K}_k \mathbf{y}_k
\end{align}
Where $\hat{\mathbf{x}}_{k|k}$ is the current best state estimate, $\mathbf{z}_k$ the measurement at the current time step, $\mathbf{R}_k$ the measurement noise covariance matrix, $h(\cdot)$ is the measurement function, which outputs the position of the obstacle, and $\mathbf{H}_k$ the measurement Jacobian.

\subsection{Learned 2D Gridmaps}
\label{subsec:gridnet}
In the following discussion, we treat dynamic objects as positive detections and static objects as negatives. A core limitation of clustering-based dynamic object detection is missed detections when in proximity to static obstacles, known as false negatives. For systems that cluster the point cloud before performing MOS, as in \cite{xu2025lv}, true positives and true negatives can be bundled into the same cluster, leading to the whole cluster being classified as static. We observe the same failure mode on methods that perform MOS before clustering as in \cite{schmid2023dynablox}. In these cases, however, the false negatives are not generated in the clustering step, but in subsequent nearest-neighbors checks intended to reduce the presence of artifacts attributed to noisy LiDAR readings. In this section, we introduce GridNet, an approach for encoding point cloud information across time and generating 2D dynamic grids that can help us recover lost detections from the aforementioned false negatives. 

\subsubsection{Architecture}
We use a PointPillars encoder~\cite{lang2019pointpillars} to project each LiDAR scan into a BEV pseudo-image and thus represent a sequence of $T$ point clouds as a BEV feature sequence. Concretely, for each scan we discretize the ground plane into vertical pillars, compute per-pillar features with a PointNet-style \cite{qi2018frustum} pillar feature network, and scatter them into a dense BEV feature map. We further compress each BEV feature map into a latent embedding using 2D convolutions, concatenate it with an MLP embedding of the vehicle’s body-frame velocity, and feed the resulting sequence to an LSTM of length $T$ to model temporal dynamics. The LSTM output is decoded with 2D convolutions into a binary BEV grid, where each cell indicates whether the corresponding 3D vertical column contains dynamic motion. Operating in BEV keeps the network lightweight and scalable to high resolutions compared to methods that learn directly in 3D (e.g., 3D convolutions on voxel grids). 
The coordinates of the 2D dynamic grid map are used to recover 3D detections in the detection fusion procedure (see \Cref{subsec:fusion}). 
\subsubsection{Loss}
The target grids are generated by projecting the position of dynamic obstacles into 2D BEV plane. Since the fraction of the image that is dynamic is substantially smaller than the fraction that is not, training a neural network to identify dynamic obstacles in the 2D grid means solving a highly unbalanced binary segmentation problem. To address this, we use a Binary Cross Entropy (BCE) loss with a Dice Score term to reward overlap between prediction and ground truth. We define our loss as,
\begin{equation}
    \mathcal{L}_{total} = w_{BCE} \mathcal{L}_{BCE} + w_{Dice} \mathcal{L}_{Dice} ,
\end{equation}
with, 
\begin{equation}
    \mathcal{L}_{BCE} = - w y \log(p) - (1-y) \log(1-p),
\end{equation}
and 
\begin{equation}
    \mathcal{L}_{Dice} = \frac{2 | Y \cap P|}{|Y| + |P|},
\end{equation}
where $y \in \{0, 1\}$ is the ground truth (whether the pixel is dynamic), $p \in [0, 1]$ is the predicted probability of the pixel being dynamic, $Y$ is the set of ground truth elements, and $P$ the set of predicted probabilities. $w$ is a weight that determines the severity of computing false positives. The weights $w_{BCE}$ and $w_{Dice}$ determine which term is favored in the total loss. The training and validation losses steadily decreased with $w_{BCE}=0.7$ and $w_{Dice}=0.3$. 

\subsection{Fusing Detections}
\label{subsec:fusion}
The detection fusion procedure is summarized in Algorithm~\ref{alg:fusion}. At each timestep, we receive the sets of all 3D, $\mathcal{S}_1,$ and 2D, $\mathcal{S}_2$, detections corresponding to that timestep as well as the set of EKF estimates, $\mathcal{K}$, computed in the previous timestep. We first process the 3D detections. For each detection, $s_1$, we iterate through the Kalman filter estimates, $\mathcal{K}$, and greedily assign to it the closest estimate under the $\ell_2$ norm. The selected estimate is then marked as assigned, and the corresponding detection is labeled with the estimate's id. Finally, the estimate is propagated into the current timestep and updated with the 3D detection's position measurement, $s_1.pos_t$. 
We process the 2D detections $\mathcal{S}_2$ analogously. For each $s_2 \in \mathcal{S}_2$, we consider only those filter estimates that did not receive an associated 3D measurement.
The missing estimates are greedily assigned a detection as in the 3D case and propagated into the current timestep. What makes this step different from the previous loop is that if a dynamic object is missed by the 3D detector, but is still indeed moving, then the 2D dynamic grid network, which has access to point cloud information over a short temporal window, can still detect the dynamic object. If the Kalman filter estimate corresponding to that object is propagated into one of the 2D grid's "dynamic columns", we can recover that estimate. Formally, we can define the set of recovered estimates, $\mathcal{K}_r$, as follows, 
\begin{equation}
 \{ \; k\in\mathcal{K} \; | \; \lVert s.pos_t - k.pos_{t-1} \lVert_2 \; < d_{min} \; \wedge \; \neg \; k.assigned \; \}
\end{equation}
Where $s\in\mathcal{S}_2$ is the 2D measurement greedily associated with $k$ and $d_{min}$ is the greedy distance threshold. We update every recovered estimate, $k\in\mathcal{K}_r$, with the $x$, $y$ coordinates from the 2D measurement,  $x_s^{(t)}$ and $y_s^{(t)}$, enhanced with the $z$ coordinate from that estimate, $z_k^{(t-1)}$, at the previous timestep. So that the measurement used in the update step becomes, 
\begin{equation}
s_2^{(t)} =
\begin{bmatrix}
x_s^{(t)} & y_s^{(t)} & z_k^{(t-1)}
\end{bmatrix}^\top
\label{eq:state}
\end{equation}

This works under the assumption that the height of the obstacle does not dramatically change between each 3D reading, which is true of most dynamic objects at $100ms$ intervals (a common LiDAR scanning period). This scheme enables us to recover 3D estimates efficiently, while maintaining near-zero false positives, and favoring 3D detections when available. 

\begin{algorithm}[t]
\caption{Detection Fusion Procedure}
\label{alg:fusion}
\begin{algorithmic}[1]
\State $\mathcal{S}_1 \gets$ Set of 3D detections  
\State $\mathcal{S}_2 \gets$ Set of 2D detections 
\State $\mathcal{K} \gets$ Set of EKF estimates 

\For{$s_1 \in \mathcal{S}_1$}
    \State $d_{min} \gets$ distance tolerance
    \For{$k \in \mathcal{K}$}
        \State $d_1 \gets$ $\lVert s_1.pos_t - k.pos_{t-1} \lVert_2$
        \If{$d_1 < d_{min}$}
            \State $d_{min} \gets d_1$
            \State $s_1.estimate\_id \gets k.id$
            \State $k.assigned \gets True$
            \State $k.predict()$ 
            \State $k.update(s_1.pos_t)$
        \EndIf  
    \EndFor 
\EndFor

\For{$s_2 \in \mathcal{S}_2$}
    \State $d_{min} \gets$ distance tolerance 
    \For{$k \in \mathcal{K}$} 
        \If {$k.assigned = False$}
            \State $d_2 \gets$ $\lVert s_2.pos_t - k.pos_{t-1} \lVert_2$
            \If{$d_2 < d_{min}$}
                \State $s_2.estimate\_id \gets k.id$
                \State $k.assigned \gets True$
                \State $s_2.pos_t.z \gets k.pos_{t-1}.z$
                \State $k.predict()$ 
                \State $k.update(s_2.pos_t)$
            \EndIf
        \EndIf 
    \EndFor
\EndFor

\State \Return $\mathcal{K}$
\end{algorithmic}
\end{algorithm}

\begin{figure}[t]
\centering
\includegraphics[width=\columnwidth,trim=150 0 150 0,clip]{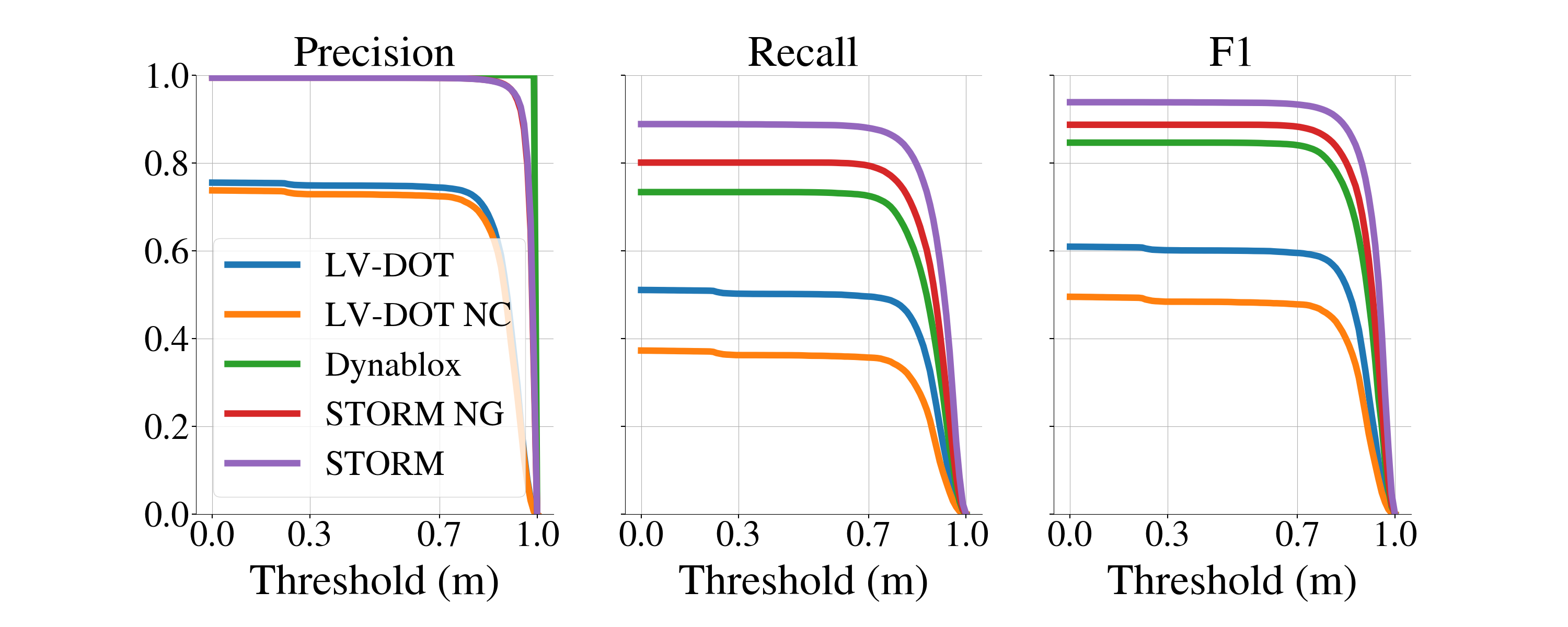}
\caption{Precision, recall, and F1 score in an experiment where a pedestrian traverses a room with 21 static obstacles. The dense obstacle configuration forces frequent strong occlusions and proximity-induced false negatives. \texttt{STORM} recovers information lost during those failure modes, achieving higher recall and F1 score than other methods.}
\label{fig:human_plot}
\vspace*{-0.1in}
\end{figure}

\begin{figure}[t]
\centering
\includegraphics[width=\columnwidth,trim=150 0 150 0,clip]{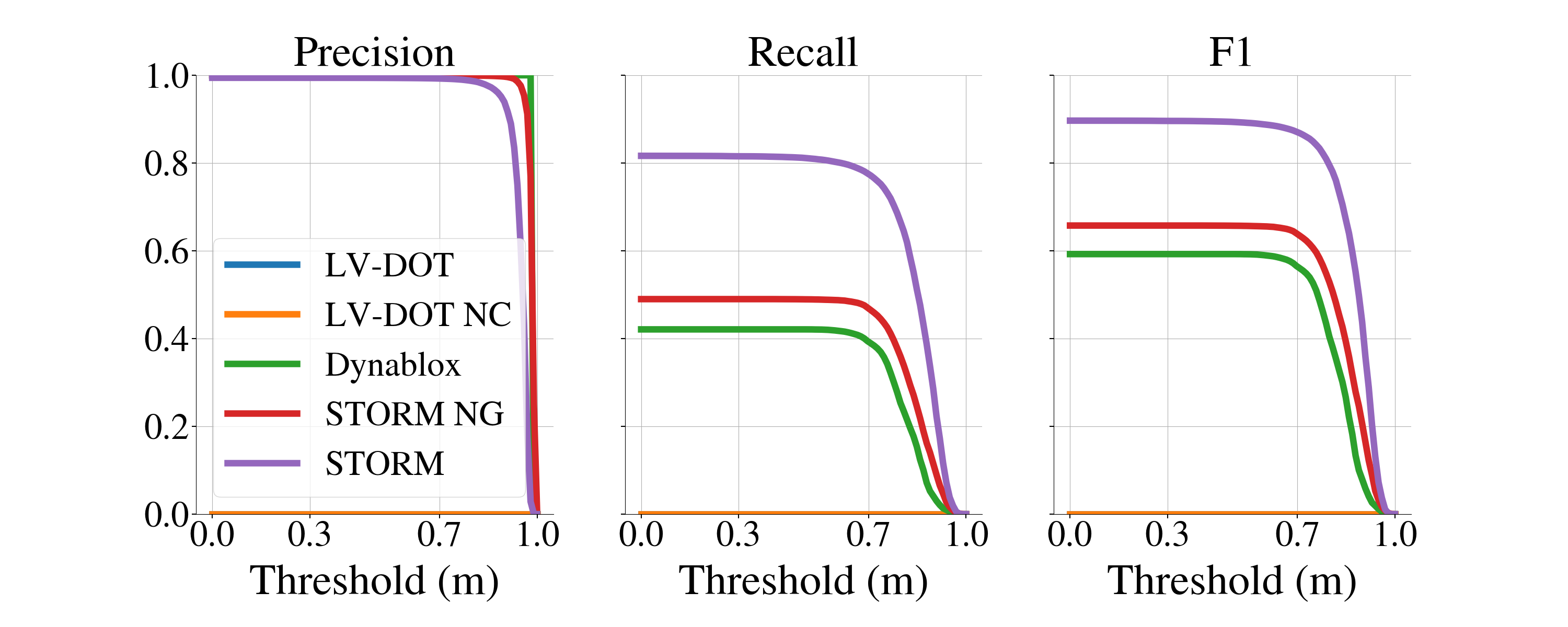}
\caption{Precision, recall, and F1 score in an experiment where a quadcopter flies at high speeds (0-5 m/s) in an empty room. \texttt{STORM} captures the high speed motion of the vehicle, achieving significantly higher recall and F1 score than other methods with comparable precision.}
\vspace*{-0.1in}
\label{fig:quad_plot}
\end{figure}

\begin{table*}[t]
\centering
\caption{Performance of methods on precision, recall, F1 score, and position error metrics across three distance thresholds on a set of experiments with a quadcopter dynamic obstacle. For each metric, the highest performing method is highlighted in \best{green} and the second highest in \secondbest{light green}. For precision, recall, and F1 score, higher is better. For position error, lower is better. \# Obs is the number of static obstacles in the room. We omit \texttt{LV-DOT}'s rows as it achieved zero precision, recall, and F1 score on these experiments.}
\label{tab:quad}
\setlength{\tabcolsep}{3pt}
\renewcommand{\arraystretch}{1.1}
\scriptsize
\resizebox{\textwidth}{!}{
\begin{tabular}{cccccccccccccc}
\toprule
\multirow{2}{*}[-0.4em]{\# Obs} & \multirow{2}{*}[-0.4em]{Method}{}
& \multicolumn{4}{c}{DistThresh = 0.75m}
& \multicolumn{4}{c}{DistThresh = 0.5m}
& \multicolumn{4}{c}{DistThresh = 0.25m} \\
\cmidrule(lr){3-6}\cmidrule(lr){7-10}\cmidrule(lr){11-14}
&
& Precision & Recall & F1 & Pos Err ($m$)
& Precision & Recall & F1 & Pos Err ($m$)
& Precision & Recall & F1 & Pos Err ($m$) \\
\midrule
\multirow{3}{*}{0}
& \texttt{Dynablox}   & \best{1.00} & 0.42 & 0.59 & \secondbest{0.21}
              & \best{1.00} & 0.42 & 0.59 & \secondbest{0.21}
              & \best{1.00} & 0.36 & 0.53 & 0.18 \\
& \texttt{STORM NG}   & \best{1.00} & \secondbest{0.49} & \secondbest{0.66} & \best{0.19}
              & \best{1.00} & \secondbest{0.49} & \secondbest{0.66} & \best{0.19}
              & \best{1.00} & \secondbest{0.44} & \secondbest{0.61} & \best{0.16} \\
& \texttt{STORM} (Ours) & \secondbest{0.99} & \best{0.82} & \best{0.90} & 0.23
              & \secondbest{0.99} & \best{0.81} & \best{0.89} & 0.22
              & 0.99 & \best{0.74} & \best{0.85} & \secondbest{0.17} \\
\midrule
\multirow{3}{*}{11}
& \texttt{Dynablox}   & \best{1.00} & \secondbest{0.50} & \secondbest{0.67} & \secondbest{0.20}
               & \best{1.00} & \secondbest{0.50} & \secondbest{0.67} & \secondbest{0.20}
               & \best{1.00} & \secondbest{0.42} & \secondbest{0.59} & \secondbest{0.16} \\
& \texttt{STORM NG}   & \best{1.00} & 0.40 & 0.57 & \best{0.18}
               & \best{1.00} & 0.40 & 0.57 & \best{0.18}
               & \best{1.00} & 0.34 & 0.51 & \best{0.15} \\
& \texttt{STORM} (Ours) & \best{1.00} & \best{0.52} & \best{0.69} & 0.21
               & \best{1.00} & \best{0.52} & \best{0.68} & 0.21
               & \best{1.00} & \best{0.43} & \best{0.61} & \secondbest{0.16} \\
\midrule
\multirow{3}{*}{26}
& \texttt{Dynablox}   & \best{1.00} & \secondbest{0.41} & \secondbest{0.58} & 0.21
               & \best{1.00} & \secondbest{0.41} & \secondbest{0.58} & 0.21
               & \best{1.00} & 0.32 & 0.48 & 0.17 \\
& \texttt{STORM NG}   & \secondbest{0.98} & 0.40 & 0.57 & \best{0.18}
               & \secondbest{0.98} & 0.40 & 0.57 & \best{0.18}
               & \secondbest{0.98} & \secondbest{0.35} & \secondbest{0.52} & \best{0.15} \\
& \texttt{STORM} (Ours) & 0.97 & \best{0.46} & \best{0.62} & \secondbest{0.20}
               & 0.97 & \best{0.45} & \best{0.62} & \secondbest{0.19}
               & 0.97 & \best{0.39} & \best{0.56} & \best{0.15} \\
\bottomrule
\end{tabular}}
\end{table*}

\section{Experiments and Results}
We evaluate our approach in two sets of experiments. The first compares the performance of state-of-the-art detection systems on human dynamic obstacles inside progressively more cluttered environments. The other compares the performance of the same methods in scenarios where the dynamic obstacle is a quadcopter. GridNet was trained on data containing no quadcopter as obstacles; consequently, no training LiDAR scans contain quadcopter returns. The success of \texttt{STORM} in those experiments indicates that the model learns motion cues, generalizing to and detecting previously unseen obstacles. Between training, validation, and testing data we ran over 40 experiments lasting between 2-10 minutes each, producing about $84,000$ labeled data points. Roughly, $57\%$ of those data points correspond to training, $7\%$ to validation, and $36\%$ to testing data.

\begin{figure}[t]
\centering
\includegraphics[width=1\columnwidth,trim=0 200 0 200,clip]{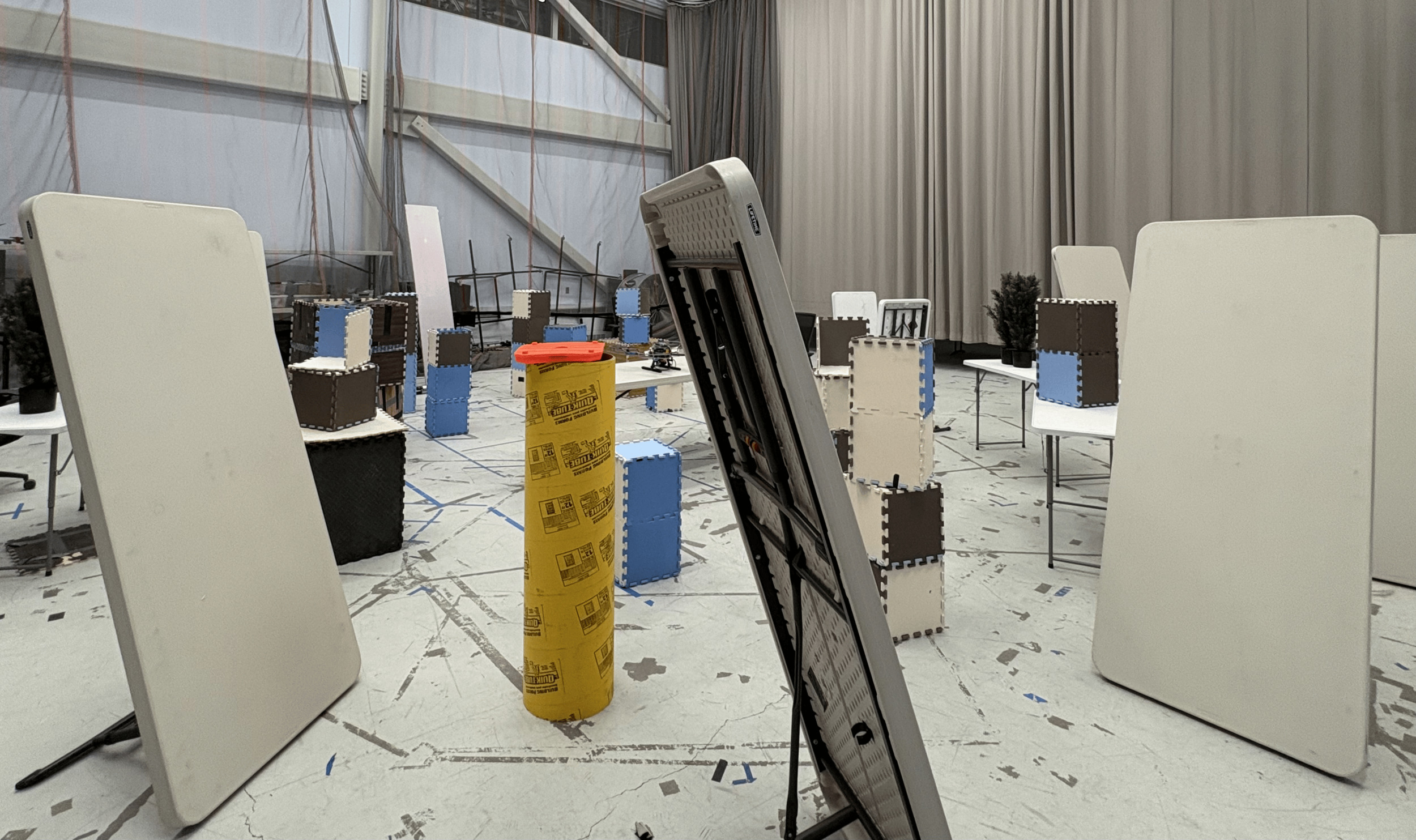}
\vspace*{-0.1in}
\caption{Experimental setup with 31 static obstacles. Quadcopter is pictured at the center of the room, surrounded by tables and foam pillars. The dense obstacle arrangement generates multiple partial occlusions and narrow corridors that clustering-based systems struggle with.}
\vspace*{-0.1in}
\label{fig:image}
\end{figure}

\subsection{Experimental Setup}
Point cloud and odometry data were collected using a Mid-360 Livox LiDAR, operated at $10~Hz$, mounted on a Holybro x500 quadcopter and a Vicon motion capture system. The Vicon motion capture system was also used to obtain the obstacle ground truth positions for each experiment. Since the motion capture system produces ground truth position as a point mass in $\mathbb{R}^3$ as opposed to 3D bounding boxes, we define the association threshold $threshold = 1 - distThresh$ as a proxy for intersection over union thresholds. Where $distThresh$ is the maximum euclidean distance for a detection to be associated to its ground truth. A high association threshold means a low maximum valid distance between detection and ground truth. 

We used collapsible tables and interlocking foam panels to build a wide range of static obstacle configurations inside of a 10x10 $m^2$ room. In between experiments, static obstacles were incrementally introduced to the space, raising the complexity of the environment and the presence of false negatives due to partial occlusions and obstacle proximity. The occupancy gridmap MOS module as well as the state estimation module were implemented in C++. The GridNet model was designed using Pytorch. We used ROS2 as the system's communication framework. All tests were run on a Lenovo Soundwave laptop with an Intel-i9 CPU and NVIDIA GEFORCE RTX 4090 GPU. For the OGM-based systems a grid resolution of $0.2~m$ and a maximum radius of $11~m$ were used throughout all experiments. The GridNet model was trained using the same dimensions and a scan history of $T=3$.

For all experiments, we compare the performance of \texttt{STORM} with \texttt{Dynablox} \cite{schmid2023dynablox} and \texttt{LV-DOT} \cite{xu2025lv}. We also include the results of running \texttt{LV-DOT} without a color image (\texttt{LV-DOT NC}) as well as \texttt{STORM} without the GridNet model (\texttt{STORM NG}) to highlight the performance benefits of the learning components of these systems. We use precision, recall, F1 score, and position error metrics, across distance thresholds ranging from $0~m$ to $1~m$, to quantify detection rate and quality of the aforementioned methods. We particularly care about the F1 score, as it captures both detection rate and quality.

\subsection{Qualitative Results and Failure Analysis}
We constructed complex static environments (e.g. \Cref{fig:image}) inside the same room to replicate the challenge of detecting dynamic obstacles in cluttered scenes in a controlled space with reliable ground truth. The arranged obstacle configurations generated strong partial occlusions and narrow gaps for dynamic objects to navigate through. While these environments exploited the limitations of \texttt{Dynablox} and \texttt{LV-DOT}, \texttt{STORM} consistently demonstrated robustness to these failure modes. Figure~\ref{fig:storm_exp} contains snapshots from three different experiments across all methods. The top row corresponds to \texttt{LV-DOT}, middle to \texttt{Dynablox}, bottom to \texttt{STORM}, and the number of static objects in each experiment increases from left to right. For clarity, the ground truth obstacle positions are marked with red rings and the bounding boxes generated by our detection system are visualized as multi-colored rectangles in each image. The second column of Figure~\ref{fig:storm_exp} highlights a common failure mode of \texttt{LV-DOT}. A pedestrian in the bottom-right corner of the image is squeezed between two foam pillars. Even though the individual is not occluded, its proximity to the pillar makes them invisible to \texttt{LV-DOT}. Similarly, on column three, a moving obstacle travels near static one, at some timestep the dynamic and static obstacles are clustered together, then the dynamic obstacle moves away, but the static one remains misclassified as dynamic for several seconds (notice the dynamic obstacles are on the opposite side of the room by the time the snapshot is taken). \texttt{Dynablox} is most affected by partial occlusions as it relies on hand-tuned parameters to tell small dynamic obstacles apart from noise. When only a fraction of the obstacle is visible, the number of dynamic points belonging to that object can fall short of a predefined threshold and the detection is missed by the system. In column one, we observe \texttt{Dynablox} as it fails to detect a pedestrian that walks behind a static obstacle, even though a fraction of the person is visible all throughout that trajectory, while \texttt{STORM} successfully detects the pedestrian during frames containing partial occlusions. Although that is \texttt{Dynablox}'s primary source of false negatives, nearest neighbor checks can subtract dynamic points from a moving obstacle when in proximity to a static one, leading to missed detections even when the obstacle is fully in the sensor's line of sight (see pedestrian in bottom right corner of column two). Our method, instead, is robust to varying dynamic obstacle size and density as it has access to a history of point cloud scans at every timestep.  

\subsection{Detector Performance on Human Obstacles}
For this set of experiments, pedestrians walked in between static obstacles as a quadcopter equipped with a LiDAR sensor was placed at the center of the room. Each pedestrian wore a hardhat with six motion capture spheres glued to its surface to generate ground truth positions. Several experiments were performed, starting with empty rooms and adding 4-6 obstacles between tests. As illustrated in Figure~\ref{fig:human_plot}, generated from an experiment containing 21 static obstacles, \texttt{STORM} performs particularly well relative to other methods in crowded rooms.

For the sake of brevity, Table~\ref{tab:human2} only contains experiments in environments cluttered with 4, 16, and 31 static obstacles with 1-3 pedestrians moving at different speeds, but the trends revealed by this table persisted across all experiments. Through all clutter levels, our method maintains precision above 0.98 while significantly improving recall relative to \texttt{Dynablox} and \texttt{LV-DOT}, particularly at higher association thresholds ($distThresh=0.25 m$). Our position error is always within $5~cm$ of the lowest error in its respective run. The recall improvement is most pronounced in high-object-density scenes, where partial occlusions and proximity-induced clustering failures are frequent. In an experiment with 31 static obstacles and two pedestrians, our approach achieved $10.18\%$ higher recall and $6.60\%$ higher F1 score than our implementation without GridNet at the $0.25~m$ distance threshold. Relative to \texttt{Dynablox}, it achieved a $28.67\%$ and $18.50\%$ improvement on recall and F1 score respectively. For reference, in the 4 obstacle case, \texttt{STORM}'s F1 score is only $2.35\%$ higher than \texttt{Dynablox}'s. The fact that in some tests, \texttt{STORM NG} achieved higher recall than \texttt{Dynablox}, may be attributed to the EKF module, which propagates position estimates between timesteps in the absence of measurements, as well as more discriminative filtering between partial occlusions and noise. Overall we found that the performance of \texttt{STORM NG} and \texttt{Dynablox} is comparable across all metrics. Out of all methods, the most impacted by clutter was \texttt{LV-DOT}. At a distance threshold of $0.25~m$, \texttt{LV-DOT}'s precision and recall sit $425.71\%$ and $297.21\%$ below \texttt{STORM}'s, respectively. This is expected as \texttt{LV-DOT} clusters objects before performing MOS, merging dynamic and static obstacles and obscuring detections.

\begin{table}[h]
\centering
\caption{Mean processing time (ms) of parallel pipelines}
\label{tab:compute}
\begin{tabular}{cccccc}
\toprule
 & Pre-process & MOS & Clustering & Fusion & Total \\
\midrule
OGM &  1.65 & 23.56 & 0.10 & 0.15 & 25.46 \\
GridNet & - & 69.31 & - & 0.15 & 69.46 \\
\bottomrule
\end{tabular}
\end{table}

\subsection{Detector Performance on Other Obstacles}
For this set of experiments, a quadcopter was flown in between static obstacles as a second quadcopter equipped with a LiDAR sensor was placed at the center of the room. Both quadcopters had motion capture spheres used to compute their ground truth. Table~\ref{tab:quad} reports the results from experiments containing 0, 11, and 26 static obstacles. 

In the most cluttered environment, \texttt{STORM} achieved $21.88\%$ higher on recall and $16.42\%$ higher on F1 score than \texttt{Dynablox}. Without GridNet, performance dropped by $11.61\%$ and $7.90\%$ on recall and F1 score respectively. Surprisingly, for these tests the gap in performance was larger when no static obstacles were present. As illustrated in Figure~\ref{fig:quad_plot}, \texttt{STORM} achieved $61.10\%$ and $39.87\%$ higher F1 score than \texttt{Dynablox} and \texttt{STORM NG}, respectively, in this experiment. While visualizing obstacle detections and ground truth positions, \texttt{STORM NG} and \texttt{Dynablox} were observed to miss detections when the obstacle accelerated. This is likely a combination of fast flight speeds, that are easily achieved in uncluttered spaces, and the quadcopter's low point density. These make OGM-based detectors struggle, as they rely on hand-tuned parameters for filtering out artifacts and sparse fast moving vehicles can be hard to distinguish from noise. Our model instead learns these parameters and performs well on low-density fast-moving obstacles as their displacement is more pronounced in the $T=3$ sized window. 

Either due to aggressive point-count, cluster-size, or cluster-density filtering catered to obstacles of human proportions, \texttt{LV-DOT}'s 3D pipeline failed to detect the dynamic quadcopter. As the quadcopter does not fit into any of YOLO's categories, it also was not detected by the vision module, leading to an overall zero precision and recall for all quadcopter tests. 

\subsection{Runtime Analysis}
\label{subsec:runtime}
Our parallel OGM-based and learning-based pipelines had a mean total runtime of $25.46~ms$ and $69.46~ms$, respectively. This corresponds to an operating frequency of $39.28~Hz$ for the OGM approach and $14.40~Hz$ for the model which is sufficient for real-time operation. Since GridNet supplements OGM detections, the model's runtime does not bottleneck the system, but in cluttered scenes, increased reliance on GridNet can make detections appear slightly delayed relative to the ground truth (see \Cref{fig:storm_exp}), reflecting GridNet’s higher inference time. This additional latency is bounded and preferable to the missed detections observed when relying on OGMs alone. While we process the point cloud histories with the PointPillars encoder sequentially, this procedure can be fully parallelized. To preserve information, we also process all $20,000$ LiDAR points per scan. We expect that parallelizing the point cloud histories, downsampling the point cloud, and exporting the Pytorch model to TensorRT should provide a performance boost sufficient to enable the deployment of our system in resource constrained embedded devices as well, such as a Jetson Orin AGX. 

\section{Conclusion and Discussion}
This paper presented \texttt{STORM}, a LiDAR-based class-agnostic multi-object detection framework robust to partial occlusions and proximity-induced failures in cluttered indoor environments. By combining temporal occupancy grid motion segmentation with a learned 2D dynamic grid representation, the proposed approach addresses key failure modes of clustering-based systems, particularly missed detections near static obstacles. 
Experimental results demonstrate consistent improvements in recall and F1 score across increasingly cluttered environments, as well as comparable precision and position error against state-of-the-art clustering-based detectors.
Finally, \texttt{STORM} demonstrated robustness to fast moving objects relative to the state-of-the-art, as it achieved the highest recall out of all methods in experiments containing obstacles moving at speeds of up to $5~m/s$.

While the OGM-GridNet fusion addresses many of the challenges faced by \texttt{Dynablox} and \texttt{LV-DOT}, we still inherit some of the limitations of OGM-based detectors. Thin surfaces generate spurious artifacts that lead to false positives, lowering overall F1 score. Moreover, since GridNet relies on a temporal window of three LiDAR scans, the model performs better on faster objects, as their displacement within the $0.3~s$ window is more pronounced. Implementing the GridNet improvements discussed in \Cref{subsec:runtime}, as well as targeted optimizations to the OGM segmentation pipeline, is a promising direction for future work as this would reduce end-to-end latency and enable efficient deployment on resource constrained embedded platforms. 





\bibliographystyle{IEEEtran}
\bibliography{references}
\end{document}

%% file: introduction.tex
Reliable detection of dynamic obstacles is crucial for safe autonomous navigation in cluttered environments. In particular, indoor scenes contain frequent close-proximity interactions between moving objects and static structure, and these interactions can produce sparse, partially occluded LiDAR returns. As a result, dynamic object detection systems must remain robust in such scenarios while maintaining real-time performance.

A widely used strategy for LiDAR-based dynamic obstacle detection builds on occupancy-grid reasoning across time: space is discretized into voxels, and objects that ``enter'' regions previously observed as free are marked as dynamic. This principle provides a strong geometric prior and can be implemented efficiently using occupancy grid maps (OGMs), followed by clustering to obtain object-level detections \cite{schmid2023dynablox}. However, in cluttered indoor environments this pipeline can suffer miss detections when dynamic objects move close to static structure. In such cases, subsequent filtering steps (e.g., nearest-neighbor consistency checks used to suppress spurious artifacts) may inadvertently remove true dynamic returns. These effects are most pronounced when the moving object is partially occluded or when only a small fraction of its surface is visible, reducing the evidence available to clustering and denoising heuristics.

This paper proposes a hybrid dynamic object detection system that couples a classical 3D temporal-occupancy pipeline with a complementary learned, class-agnostic dynamic prior in a lower-dimensional bird's-eye-view (BEV) representation. Our approach runs two detection pipelines in parallel. The first pipeline, inspired by Dynablox \cite{schmid2023dynablox}, performs moving object segmentation in a temporal OGM and clusters dynamic voxels to produce 3D detections, which are subsequently tracked using an Extended Kalman Filter (EKF) \cite{kalman1960new}. The second pipeline aggregates a short history of LiDAR scans (and corresponding vehicle states), encodes them using the PointPillars encoder \cite{lang2019pointpillars}, and models temporal context using a Long Short Term Memory (LSTM) module. The network outputs a 2D dynamic grid in which each cell represents whether the corresponding vertical column in 3D contains dynamic motion. Importantly, this representation is class-agnostic: rather than recognizing a closed set of semantic categories, it learns to predict dynamic regions from short-term spatiotemporal cues.

To exploit the complementary strengths of these two pipelines, we introduce $\textbf{STORM}$ ($\textbf{S}$patio$\textbf{T}$emporal-cues and $\textbf{O}$ccupancy-based $\textbf{R}$obust $\textbf{M}$otion detection), a detection fusion procedure that prioritizes 3D OGM-based detections when available, while using 2D dynamic-grid evidence to recover tracks that would otherwise be lost due to proximity-induced false negatives (see Figure~\ref{fig:storm_exp}). 

We evaluate the proposed system on real-world experiments in cluttered indoor environments and show that the learned dynamic prior significantly reduces false negatives caused by dynamic-static proximity. 
We additionally compare against a LiDAR--visual baseline to highlight the limitations of vision-based approaches in cluttered indoor scenes, including restricted camera field-of-view and closed-set semantics.

In summary, our contributions are:
\begin{itemize}
    \item a spatiotemporal network that predicts class-agnostic BEV dynamic grids from short LiDAR histories and ego-motion;
    \item a fusion strategy that uses the learned dynamic grid to recover missed 3D detections and maintain tracks during proximity and occlusion-induced failures;
    \item extensive validation with real-world data on indoor scenes with complex static obstacle configurations and multiple dynamic obstacle classes reporting higher recall and F1 score than the state-of-the-art.
\end{itemize}

%% file: related_work.tex
\begin{figure*}[th]
\centering
\includegraphics[width=\textwidth]{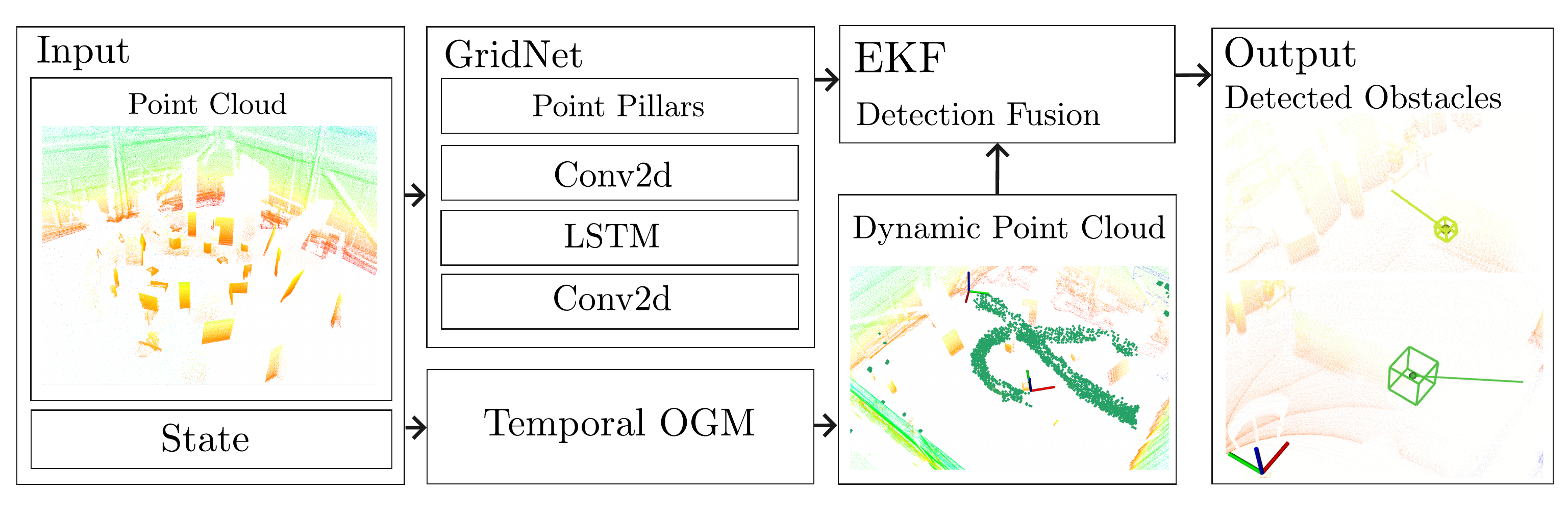}
\caption{System architecture of the proposed detection framework: point cloud and ego state information are processed in parallel by our OGM and learning-based MOS pipelines. The cluster centroids of the dynamic point cloud produced by the OGM module and GridNet's 2D dynamic grids are fused to generate dynamic obstacle detections.}
\label{fig:system}
\end{figure*}

\subsection{LiDAR-based Dynamic Object Detection}
A common approach to LiDAR-based dynamic object detection \cite{min2021kernel,schreiber2021dynamic,lu2024fapp, shi2020pv, wu2024moving, yin2021center} is to reason about motion through temporal consistency in an occupancy representation~\cite{elfes2002using}. 
Within this family, Dynablox~\cite{schmid2023dynablox} is a representative work that performs moving object segmentation by explicitly reasoning over temporal occupancy and free-space consistency, and then applies clustering to obtain object-level detections.
While effective in outdoor autonomous driving scenarios, these pipelines can suffer false negatives in cluttered indoor scenes when dynamic objects move near static structure or are only partially observed: dynamic and static returns can spatially overlap, and subsequent clustering or denoising heuristics can suppress the sparse traces of true motion. This motivates complementary mechanisms that aggregate short-term temporal cues and provide a prior for regions where instantaneous geometric cues are weak.

In addition to LiDAR-only pipelines, vision-augmented methods leverage semantic cues to improve detection of dynamic objects from background. However, these approaches are often constrained by closed-set labels and by the camera field-of-view (FOV), and can therefore miss novel obstacles or objects outside the image frustum. 
For example, LV-DOT~\cite{xu2025lv} uses an RGB-D camera to provide additional visual cues, including an RGB detector based on a pretrained closed-set model (i.e., YOLO \cite{redmon2016you}) that identifies dynamic obstacles from predefined classes~\cite{xu2025lv}. In addition, because the visual cues are only available within the camera frustum, LV-DOT may fail when obstacles are outside the camera FOV, even if they are observed by LiDAR.

\subsection{Learning-based Moving Object Segmentation}
Learning-based moving object segmentation (MOS) labels observations as moving versus static by leveraging spatiotemporal cues from sequential LiDAR data, typically without requiring explicit semantic categories \cite{hoermann2018dynamic,chen2021moving,mersch2022receding,mersch2023building,zhou2023motionbev}. Chen \etal~\cite{chen2021moving} show that exploiting temporal information in 3D LiDAR sequences can substantially improve MOS compared to single-scan reasoning. Subsequent work has advanced both representations and learning strategies for MOS, including sparse 4D convolutions for spatiotemporal processing~\cite{mersch2022receding}, map-based MOS that builds volumetric beliefs in dynamic environments~\cite{mersch2023building}, and automatic labeling procedures for scalable online MOS training data generation~\cite{chen2022automatic}.
BEV-centric online MOS methods have also been proposed, combining appearance and motion cues in a BEV representation for efficiency~\cite{zhou2023motionbev}.

%% file: references.bib
@article{schmid2023dynablox,
  title={Dynablox: Real-time detection of diverse dynamic objects in complex environments},
  author={Schmid, Lukas and Andersson, Olov and Sulser, Aurelio and Pfreundschuh, Patrick and Siegwart, Roland},
  journal={IEEE Robotics and Automation Letters},
  volume={8},
  number={10},
  pages={6259--6266},
  year={2023},
  publisher={IEEE}
}

@inproceedings{lang2019pointpillars,
  title={{Pointpillars}: Fast encoders for object detection from point clouds},
  author={Lang, Alex H and Vora, Sourabh and Caesar, Holger and Zhou, Lubing and Yang, Jiong and Beijbom, Oscar},
  booktitle={Proceedings of the IEEE/CVF conference on computer vision and pattern recognition},
  pages={12697--12705},
  year={2019}
}

@article{xu2025lv,
  title={{LV-DOT}: LiDAR-visual dynamic obstacle detection and tracking for autonomous robot navigation},
  author={Xu, Zhefan and Shen, Haoyu and Han, Xinming and Jin, Hanyu and Ye, Kanlong and Shimada, Kenji},
  journal={arXiv preprint arXiv:2502.20607},
  year={2025}
}

@article{mersch2023building,
  title={Building volumetric beliefs for dynamic environments exploiting map-based moving object segmentation},
  author={Mersch, Benedikt and Guadagnino, Tiziano and Chen, Xieyuanli and Vizzo, Ignacio and Behley, Jens and Stachniss, Cyrill},
  journal={IEEE Robotics and Automation Letters},
  volume={8},
  number={8},
  pages={5180--5187},
  year={2023},
  publisher={IEEE}
}

@inproceedings{redmon2016you,
  title={You only look once: Unified, real-time object detection},
  author={Redmon, Joseph and Divvala, Santosh and Girshick, Ross and Farhadi, Ali},
  booktitle={Proceedings of the IEEE conference on computer vision and pattern recognition},
  pages={779--788},
  year={2016}
}

@article{lu2024fapp,
  title={Fapp: Fast and adaptive perception and planning for uavs in dynamic cluttered environments},
  author={Lu, Minghao and Fan, Xiyu and Chen, Han and Lu, Peng},
  journal={IEEE Transactions on Robotics},
  year={2024},
  publisher={IEEE}
}

@inproceedings{qi2018frustum,
  title={Frustum pointnets for 3d object detection from rgb-d data},
  author={Qi, Charles R and Liu, Wei and Wu, Chenxia and Su, Hao and Guibas, Leonidas J},
  booktitle={Proceedings of the IEEE conference on computer vision and pattern recognition},
  pages={918--927},
  year={2018}
}

@article{elfes2002using,
  title={Using occupancy grids for mobile robot perception and navigation},
  author={Elfes, Alberto},
  journal={Computer},
  volume={22},
  number={6},
  pages={46--57},
  year={2002},
  publisher={IEEE}
}

@inproceedings{yin2021center,
  title={Center-based 3d object detection and tracking},
  author={Yin, Tianwei and Zhou, Xingyi and Krahenbuhl, Philipp},
  booktitle={Proceedings of the IEEE/CVF conference on computer vision and pattern recognition},
  pages={11784--11793},
  year={2021}
}

@article{zhou2023motionbev,
  title={{MotionBEV}: Attention-aware online LiDAR moving object segmentation with bird's eye view based appearance and motion features},
  author={Zhou, Bo and Xie, Jiapeng and Pan, Yan and Wu, Jiajie and Lu, Chuanzhao},
  journal={IEEE Robotics and Automation Letters},
  volume={8},
  number={12},
  pages={8074--8081},
  year={2023},
  publisher={IEEE}
}

@article{chen2021moving,
  title={Moving object segmentation in 3D LiDAR data: A learning-based approach exploiting sequential data},
  author={Chen, Xieyuanli and Li, Shijie and Mersch, Benedikt and Wiesmann, Louis and Gall, J{\"u}rgen and Behley, Jens and Stachniss, Cyrill},
  journal={IEEE Robotics and Automation Letters},
  volume={6},
  number={4},
  pages={6529--6536},
  year={2021},
  publisher={IEEE}
}

@article{mersch2022receding,
  title={Receding moving object segmentation in 3d lidar data using sparse 4d convolutions},
  author={Mersch, Benedikt and Chen, Xieyuanli and Vizzo, Ignacio and Nunes, Lucas and Behley, Jens and Stachniss, Cyrill},
  journal={IEEE Robotics and Automation Letters},
  volume={7},
  number={3},
  pages={7503--7510},
  year={2022},
  publisher={IEEE}
}

@article{chen2022automatic,
  title={Automatic labeling to generate training data for online LiDAR-based moving object segmentation},
  author={Chen, Xieyuanli and Mersch, Benedikt and Nunes, Lucas and Marcuzzi, Rodrigo and Vizzo, Ignacio and Behley, Jens and Stachniss, Cyrill},
  journal={IEEE Robotics and Automation Letters},
  volume={7},
  number={3},
  pages={6107--6114},
  year={2022},
  publisher={IEEE}
}

@article{kalman1960new,
  title={A new approach to linear filtering and prediction problems},
  author={Kalman, Rudolph Emil},
  year={1960}
}

@inproceedings{min2021kernel,
  title={Kernel-based 3-D dynamic occupancy mapping with particle tracking},
  author={Min, Youngjae and Kim, Do-Un and Choi, Han-Lim},
  booktitle={2021 IEEE International Conference on Robotics and Automation (ICRA)},
  pages={5268--5274},
  year={2021},
  organization={IEEE}
}

@inproceedings{schreiber2021dynamic,
  title={Dynamic occupancy grid mapping with recurrent neural networks},
  author={Schreiber, Marcel and Belagiannis, Vasileios and Gl{\"a}ser, Claudius and Dietmayer, Klaus},
  booktitle={2021 IEEE International Conference on Robotics and Automation (ICRA)},
  pages={6717--6724},
  year={2021},
  organization={IEEE}
}

@inproceedings{hoermann2018dynamic,
  title={Dynamic occupancy grid prediction for urban autonomous driving: A deep learning approach with fully automatic labeling},
  author={Hoermann, Stefan and Bach, Martin and Dietmayer, Klaus},
  booktitle={2018 IEEE International Conference on Robotics and Automation (ICRA)},
  pages={2056--2063},
  year={2018},
  organization={IEEE}
}

@article{wu2024moving,
  title={Moving event detection from LiDAR point streams},
  author={Wu, Huajie and Li, Yihang and Xu, Wei and Kong, Fanze and Zhang, Fu},
  journal={nature communications},
  volume={15},
  number={1},
  pages={345},
  year={2024},
  publisher={Nature Publishing Group UK London}
}

@inproceedings{shi2020pv,
  title={Pv-rcnn: Point-voxel feature set abstraction for 3d object detection},
  author={Shi, Shaoshuai and Guo, Chaoxu and Jiang, Li and Wang, Zhe and Shi, Jianping and Wang, Xiaogang and Li, Hongsheng},
  booktitle={Proceedings of the IEEE/CVF conference on computer vision and pattern recognition},
  pages={10529--10538},
  year={2020}
}
